\documentclass[sigconf]{acmart}
\usepackage{multirow}
\usepackage{multicol}
\usepackage{enumitem}
\usepackage{subfigure}
\usepackage{amsfonts}
\usepackage{amsmath}
\usepackage{bbm}
\usepackage{xspace}
\usepackage{algpseudocode}
\usepackage[ruled,vlined,linesnumbered]{algorithm2e}
\usepackage{xcolor}
\usepackage{bbding}

\copyrightyear{2023}
\acmYear{2023}
\setcopyright{rightsretained}
\acmConference[KDD '23]{Proceedings of the 29th ACM SIGKDD Conference on Knowledge Discovery and Data Mining}{August 6--10, 2023}{Long Beach, CA, USA}
\acmBooktitle{Proceedings of the 29th ACM SIGKDD Conference on Knowledge Discovery and Data Mining (KDD '23), August 6--10, 2023, Long Beach, CA, USA}
\acmDOI{10.1145/3580305.3599443}
\acmISBN{979-8-4007-0103-0/23/08}

\newcommand{\method}[1]{\mbox{\sf #1}\xspace}
\newcommand{\our}{\method{\textsc{FedAlign}}}
\newcommand{\ourbold}{\method{\textsc{\textbf{FedAlign}}}}
\newcommand{\smallsection}[1]{\noindent\textbf{#1}.}
\newtheorem{theorem}{Theorem}[section]
\newtheorem{corollary}{Corollary}[theorem]

\begin{document}

\widowpenalty=10

\title{Navigating Alignment for Non-identical Client Class Sets: A Label Name-Anchored Federated Learning Framework}

\author{Jiayun Zhang}
\affiliation{
  \institution{University of California, San Diego}
  \country{}
  }
\email{jiz069@ucsd.edu}

\author{Xiyuan Zhang}
\affiliation{
  \institution{University of California, San Diego}
  \country{}
  }
\email{xiyuanzh@ucsd.edu}

\author{Xinyang Zhang}
\affiliation{
  \institution{University of Illinois at Urbana-Champaign}
  \country{}
  }
\email{xz43@illinois.edu}

\author{Dezhi Hong}
\authornote{Work unrelated to Amazon.}
\affiliation{
  \institution{Amazon}
  \country{}
  }
\email{hondezhi@amazon.com}

\author{Rajesh K. Gupta}
\affiliation{
  \institution{University of California, San Diego}
  \country{}
  }
\email{rgupta@ucsd.edu}

\author{Jingbo Shang}
\affiliation{
  \institution{University of California, San Diego}
  \country{}
  }
\email{jshang@ucsd.edu}

\renewcommand{\shortauthors}{Jiayun Zhang et al.}

\begin{abstract}
    Traditional federated classification methods, even those designed for non-IID clients, assume that each client annotates its local data with respect to the same universal class set. In this paper, we focus on a more general yet practical setting, \emph{non-identical client class sets}, where clients focus on their own (different or even non-overlapping) class sets and seek a global model that works for the union of these classes. If one views classification as finding the best match between representations produced by data/label encoder, such heterogeneity in client class sets poses a new significant challenge---local encoders at different clients may operate in different and even independent latent spaces, making it hard to aggregate at the server. We propose a novel framework, \our\footnote{Code is available on GitHub: \url{https://github.com/jiayunz/FedAlign}.}, to align the latent spaces across clients from both label and data perspectives. From a label perspective, we leverage the expressive natural language class names as a common ground for label encoders to anchor class representations and guide the data encoder learning across clients. From a data perspective, during local training, we regard the global class representations as anchors and leverage the data points that are close/far enough to the anchors of locally-unaware classes to align the data encoders across clients. Our theoretical analysis of the generalization performance and extensive experiments on four real-world datasets of different tasks confirm that \our outperforms various state-of-the-art (non-IID) federated classification methods.
\end{abstract}



\maketitle

\section{Introduction}

Federated learning~\cite{mcmahan2017communication} has emerged as a distributed learning paradigm that allows multiple parties to collaboratively learn a global model effective for all participants
while preserving the privacy of their local data.
It brings benefits to various domains, such as recommendation systems~\cite{liang2021fedrec++,liu2021fedct,yi2021efficient}, ubiquitous sensing~\cite{li2021meta,tu2021feddl,li2021fedmask} and mobile computing~\cite{yang2021characterizing,li2021hermes,li2021talk}.

\begin{figure}[t]
    \centering
    \subfigure[The problem setting of non-identical client class sets.]{
    \centering
    \includegraphics[width=0.66\linewidth]{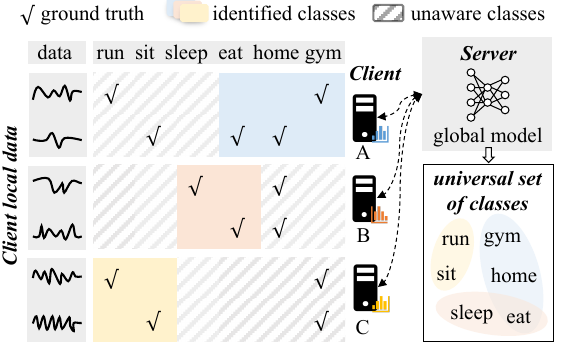}
    \label{fig:intro-problem-setting}
    }
    \subfigure[Unique challenge.]{
    \centering
    \includegraphics[width=0.28\linewidth]{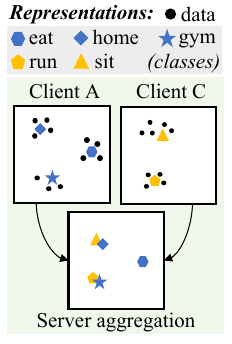}
    \label{fig:intro-matching}
    }
    \caption{
    Illustrations of our problem setting and unique challenge of misaligned latent spaces across clients, using a behavioral context recognition system where users have different preferences in reporting (i.e., annotating) labels.
    }
\end{figure}

\begin{figure*}[t]
    \centering
    \includegraphics[width=\linewidth]{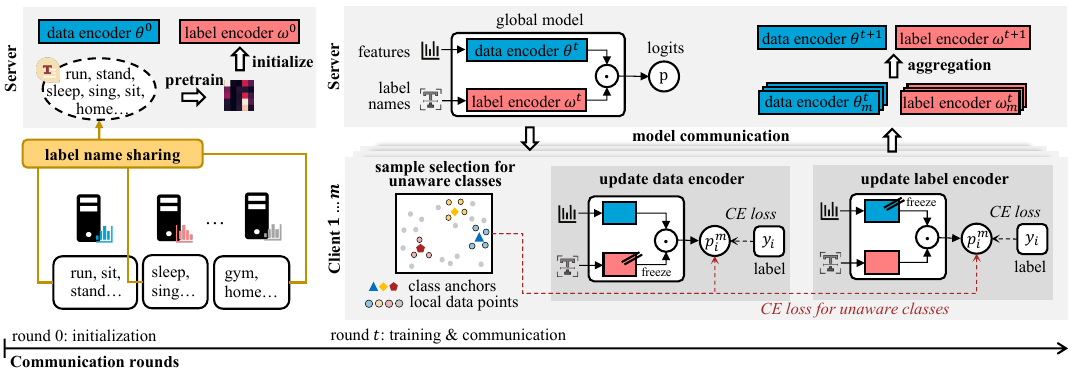}
    \caption{
    Overview of \our framework. The label names are leveraged as a common ground for label encoders to anchor class representations. During local training, the two encoders perform alternating training to mutually regulate the latent spaces. The global class representations are regarded as class anchors. Pseudo-labels are assigned to partially-unlabeled local samples for unaware classes based on their distances to the anchors. An additional cross-entropy loss for unaware classes is added to the local learning objective to reduce the divergence between global and local distributions. 
    }
    \label{fig:method-overview}
\end{figure*}

Existing federated classification methods~\cite{li2020federated,li2021model,karimireddy2020scaffold,wang2020federated,lin2020ensemble,wang2020tackling,zhu2021data,luo2021no} typically assume that the local annotations at each client follow the same set of classes; however, this assumption does not hold true in many real-world applications.
For example, a smartwatch company wants to build a human activity classifier for all activity types, as shown in Figure~\ref{fig:intro-problem-setting}. Although their smartwatch users as clients could experience almost all types of daily activities, each user may only opt to report (i.e., annotate) a subset of activities.
Another example is a federated medical diagnosis system, which attempts to infer all types of diseases of a patient for comprehensive health screening. Physicians and specialist groups with different expertise can participate in this federated learning system as clients. As one can see here, different specialists will only offer disease annotations within their domains, even if a patient may have several types of diseases at the same time. This makes the class sets at many clients non-identical and even non-overlapping.

We aim to lift this assumption and work on a general and rather practical federated learning setting, \textbf{non-identical client class sets}, where clients focus on their own (different or even non-overlapping) class sets and seek a global classification model that works for the union of these classes. 
We denote the classes that are not covered in the local annotations as \emph{locally-unaware classes}. 
Note that each client can have local data whose true labels are among the locally-unaware classes.
Also, the classification task here can be either single-label or multi-label. When it is multi-label, the local data might be only partially labeled due to the locally-unaware classes.
Therefore, this new setting is more general and challenging than the missing class scenario~\cite{li2021fedrs} which assumes the single-label scenario and no local data is from locally-unaware classes. 

The non-identical client class sets pose a significant challenge of huge variance in local training across different clients. 
As shown in Figure~\ref{fig:intro-matching}, one can view classification as a matching process between data representations and label representations in a latent space.
Because of the non-identical client class sets, locally trained classifiers are more likely to operate in drastically different latent spaces.
Moreover, when the class sets are non-overlapping, it is possible that the latent spaces at different clients are completely independent. 
This would result in inaccurate classification boundaries after aggregation at the server, making our setting more challenging than non-IID clients with identical client class sets.

We propose a novel federated learning framework \our, as shown in Figure~\ref{fig:method-overview}, to align the latent spaces across clients from both label and data perspectives as follows:
\begin{enumerate}[nosep,leftmargin=*]
    \item[(1)] \emph{Anchor the label representations using label names}. 
        We observe that the natural-language class names (i.e., label names) often carry valuable information for understanding label semantics, and more importantly, they are typically safe to share with all parties. Therefore, we break the classification model into a data encoder and a label encoder as shown in Figure~\ref{fig:method-overview}, and then leverage the label names as the common ground for label encoders. The server initializes the label encoder with pretrained text representations, such as word embedding. The label encoder will be then distributed to different clients and updated alternatingly with data encoders during local training and global aggregation, mutually regulating the latent space.
    \item[(2)] \emph{Connect the data representations via anchors of locally-unaware classes}.
        During local training, we regard the global class representations as anchors and utilize data points that are close/far enough to the anchors of locally-unaware classes to align the data encoders. Specifically, as shown in Figure~\ref{fig:method-overview}, at each client, we annotate local data based on their distances to the anchors and add another cross-entropy loss between the pseudo-labels and the model predictions. Such regularization encourages the data encoders to reside in the same latent space.
\end{enumerate}

Our theoretical analysis shows that \our can achieve a better generalization bound than traditional federated learning methods, suggesting a strong potential for performance improvement.
Experiments on four real-world datasets, including the most challenging scenario of multi-label classification and non-overlapping client class sets, confirm that \our outperforms various state-of-the-art (non-IID) federated classification methods.
Our contributions are summarized as follows:
\begin{itemize}[leftmargin=*,nosep]
    \item We propose a more general yet practical federated classification setting, namely non-identical client class sets.
    We identify the new challenge caused by the heterogeneity in client class sets --- local models at different clients may operate in different and even independent latent spaces.
    \item We propose a novel framework \our to align the latent spaces across clients from both label and data perspectives.
    \item Our generalization bound analysis and extensive experiments on four real-world datasets of different tasks confirm the superiority of \our over various state-of-the-art (non-IID) federated classification methods both theoretically and empirically.
\end{itemize}
\section{Preliminaries} \label{sec:preliminary}
\smallsection{Problem Formulation}
We aim to generate a global classification model using federated learning with non-identical class sets, where each client only identifies part of the classes from its dataset. Denote the universal set of classes as $\mathbf{C}$, the set of classes that are identified on client $m$ is $\mathbf{C}_m$, and the set of locally-unaware classes is $\overline{\mathbf{C}_m}$, where $\mathbf{C}_m \cup \overline{\mathbf{C}_m} = \mathbf{C}$. The goal is to learn a global model $g: \mathcal{X} \rightarrow \{0, 1\}^{\lvert \mathbf{C}\rvert}$ that given $x \in \mathcal{X}$, all positive labels from $\mathbf{C}$ can be inferred.

The training set on client $m$ is denoted as $\mathbf{D}_m = \{(x_i, y_i)\}_{i=1}^{N}$, where $x_i$ is the input data and $y_i = [y_{i, c}]_{c \in \mathbf{C}}$ is a vector showing the labels of each class. If $c \in \mathbf{C}_m$, $y_{i, c} \in \{0, 1\}$. If $c \in \overline{\mathbf{C}_m}$, $y_{i, c}$ is unknown. It is possible that some data samples $x_i\in \mathbf{D}_m$ do not belong to any of the classes in $\mathbf{C}_m$, i.e., $\forall c \in \mathbf{C}_m: y_{i,c} = 0$.

\smallsection{Backbone Classification Model}
Let $\mathcal{Z} \subset \mathcal{R}^d$ be the latent feature space and $\mathcal{Y}$ be the output spaces. Generally, the classification model $g$ can be decomposed into a data encoder $f : \mathcal{X} \rightarrow \mathcal{Z}$ parameterized by $\theta$ and a linear layer (i.e., classifier) $h : \mathcal{Z} \rightarrow \mathcal{Y}$ parameterized by $\zeta$. The data encoder $f$ generates representations for input data. Then, the classifier $h$ transforms the representations into prediction logits. Given an input $x_i$, the predicted probability given by $g$ is $g(x_i;\theta,\zeta) = \sigma(h(f(x_i;\theta);\zeta))$, where $\sigma$ is the activation function. We discuss two types of classification tasks as follows. 

\smallsection{Single-Label Multi-Class Classification}
In this setting, each sample is associated with only one positive class. In other words, the classes are mutually exclusive. We use softmax activation to get the predicted probability. The class with the maximum probability is predicted as the positive class. Let $g(x_i;\theta,\zeta)_c$ denote the predicted probability of $x_i$ belonging to class $c$. During training, the cross-entropy loss is used as the loss function:
\begin{equation}
    \ell(g(x_i;\theta,\zeta), y_i) = -\sum\nolimits_{c\in \mathbf{C}_m}y_{i,c}\log g(x_i; \theta,\zeta)_c.
    \label{eq:cross-entropy-loss}
\end{equation}

\smallsection{Multi-Label Classification}
In this setting, each sample may be associated with a set of positive classes. For example, a person may have both diabetes and hypertension. The sigmoid activation is applied to get the predicted probability. Each element in the predicted probability represents the probability that the input data $x_i$ is associated with a specific class. The final predictions are achieved by thresholding the probabilities at 0.5. If $g(x_i;\theta,\zeta)_c > 0.5$, $x_i$ is predicted to be associated with class $c$. During training, the binary cross-entropy loss is used as the loss function:
\begin{equation}
\begin{aligned}
    \ell(g(x_i;\theta,\zeta), y_i) = -\sum\nolimits&_{c\in \mathbf{C}_m}[y_{i,c}\log g(x_i;\theta,\zeta)_c \\
    &+ (1-y_{i,c})\log(1-g(x_i;\theta,\zeta)_c)].
\end{aligned}
\label{eq:binary-cross-entropy-loss}
\end{equation}

\smallsection{Federated Learning}
Consider a federated learning system with $M$ clients. The server coordinates $M$ clients to update the model in $T$ communication rounds. The learning objective is to minimize the loss on every client, i.e., $\mathop{\text{min}}_{\theta,\zeta} \frac{1}{M}\sum_{m\in [M]} \mathcal{L}_{m}(\theta,\zeta)$. At each round, the server sends the model parameters to a subset of clients and lets them optimize the model by minimizing the loss over their local datasets. The loss at client $m$ is:
\begin{equation}
    \mathcal{L}_{m}(\theta,\zeta) = \mathbb{E}_{(x_i, y_i)\sim \mathbf{D}_m}\ell(g(x_i;\theta,\zeta), y_i).
    \label{eq:general-local-objective}
\end{equation}

At the end of each round, the server aggregates the model parameters received from clients, usually by taking the average.
\section{The \our Framework}\label{sec:method}

\begin{algorithm}[htbp]
    \SetKwData{Left}{left}\SetKwData{This}{this}\SetKwData{Up}{up}
    \SetKwFunction{Union}{Union}\SetKwFunction{FindCompress}{FindCompress}
    \SetKwInOut{Input}{Input}\SetKwInOut{Output}{Output}
    \Input{Communication rounds $T$, number of selected clients per round $\lvert \mathbf{S}_t\rvert$, local training epochs $E$.}
    \Output{The final global model $g(x;\theta^T, \omega^T)$.}

    \textbf{\underline{Server executes}:}\\
    Collect label names from clients and pretrain text representations to initialize label encoder $\omega^0$\; 
    Randomly initialize data encoder $\theta^0$\;
    \For{$t = 0, 1, \dots, T-1$}{
        Select a subset $\mathbf{S}_t$ of clients at random\; 
        \For{$m \in \mathbf{S}_t$}{
            $\theta^{(t+1)}_{m}, \omega^{(t+1)}_{m} \leftarrow \textbf{ClientUpdate}(m, \theta^{(t)}, \omega^{(t)})$\;
        }
        $\theta^{(t+1)} \leftarrow \frac{\sum\nolimits_{m\in \mathbf{S}_t}\theta_{m}^{(t+1)}}{\lvert \mathbf{S}_t\rvert}$;
        $\omega_c^{(t+1)} \leftarrow \frac{\sum\nolimits_{m\in \mathbf{S}_t, c\in \mathbf{C}_m}\omega_{m,c}^{(t+1)}}{\lvert \{m\vert m\in \mathbf{S}_t, c\in \mathbf{C}_m\}\rvert}$\;
    }
    \Return $\theta^T, \omega^T$\;
    $\textbf{ClientUpdate}(m, \theta^{(t)}, \omega^{(t)})$:\\
    $\theta^{(t)}_{m}, \omega^{(t)}_{m} \leftarrow \theta^{(t)}, \omega^{(t)}$\;
    \For{$e = 1, 2, \dots, E$}{
        Calculate distances of data points and class anchors and form dataset $D'^{(t)}_{m}$ \;
        Alternatingly update $\theta_{m}^{(t+1)}$ and $\omega_{m}^{(t+1)}$ \;
    }
    \Return $\theta^{(t+1)}_{m}$, and $\omega^{(t+1)}_{m}$ to server \;

  \caption{\our Framework}
  \label{algo:federated-framework}
\end{algorithm}

\subsection{Overview}
The pseudo code of \our can be found in Algorithm~\ref{algo:federated-framework}. Learning with \our framework consists of the following steps:
\begin{enumerate}[leftmargin=*]
    \item \textbf{Label name sharing and label encoder initialization:} 
    Before training, the server collects the natural language label names from the clients. 
    The server initializes the label encoder's parameters $\omega^0$ via pretrained text representations, such as word embedding. We expect more advanced techniques like pretrained neural language models could make the learning converge even faster, but we leave it as future work. 
    \item \textbf{Client selection and model communication:} At $t$-th round, the server randomly selects a subset of clients $\mathbf{S}_t$ and sends the global model parameters to them.
    \item \textbf{Local training:} Client $m \in \mathbf{S}_t$ independently trains its local model and returns the model parameters.
    \item \textbf{Model aggregation:} The server aggregates the parameters of client models into global parameters.
\end{enumerate}
Pretraining text representations and label encoder initialization in (1) are conducted only once at the beginning. Steps (2)-(4) repeat for $T$ rounds until the global model converges. During local training in (3), each client $m \in \mathbf{S}_t$ conducts the following steps:
\begin{enumerate}[label=(\alph*),leftmargin=*]
    \item \textbf{Select samples for unaware classes via class anchors:} Client $m$ forms a dataset $\mathbf{D}'^{(t)}_{m}$ for locally-unaware classes $\overline{\mathbf{C}_m}$ by using the latest class representations as anchors and computing the distances to the data representations. 
    \item \textbf{Alternating training of two encoders:} Client $m$ freezes the label encoder and updates the data encoder. Then, it freezes the data encoder and updates the label encoder.
    \item \textbf{Model communication after local updates:} Client $m$ sends the updated model parameters to the server.
\end{enumerate}

\subsection{Label Name-Anchored Matching}
\label{sec:method-semantic}
The vanilla model described in Section~\ref{sec:preliminary} learns feature spaces merely based on local training data with numerical label IDs. However, with non-identical client class sets, local models at different clients are likely to form different and even independent feature spaces, making the classification boundaries aggregated at the server inaccurate. To better align the feature spaces, we leverage the semantics of label names as a common reference to anchor class representations. The natural language label names carry valuable information for understanding label correlations. For example, in behavioral context recognition, the activity of ``lying down'' is likely to indicate the person is ``sleeping'', and the possible location of the activity is ``at home''. Such knowledge about label correlations not only exists in the datasets to investigate, but can also be mined through analyzing the semantics of label names. 

\smallsection{Incorporating Label Encoder to Classification Model}
We replace the classifier in a conventional classification model with a label encoder as shown in Figure~\ref{fig:method-overview}. Let $\mathcal{W}$ be the set of natural language label names with respect to $\mathbf{C}$, and $\mathcal{Z}$ be the latent feature space. The new classification model $g = f\circ \gamma$ consists of two branches: a data encoder $f: \mathcal{X} \rightarrow \mathcal{Z}$ parameterized by $\theta$ and a label encoder $\gamma : \mathcal{W} \rightarrow \mathcal{Z}$ parameterized by $\omega$. The $\circ$ is the operation to get dot product. The label encoder takes the label names $w_c \in \mathcal{W}$ as inputs and maps them into representations $\gamma(w_c;\omega)$. Prior knowledge about label semantics can be inserted into the label encoder by initializing it with pretrained label embeddings. Inspired by existing works that learn semantic word embeddings based on word-word co-occurrence~\cite{bullinaria2007extracting} and point-wise mutual information (PMI)~\cite{pennington2014glove,levy2014linguistic}, we use an external text corpus related to the domain of the classification task to extract knowledge of label co-occurrence and pretrain label embeddings for initializing the label encoder. The pretraining details can be found in Appendix.

\smallsection{Representation Matching} Given an input $x_i$, the model uses the data encoder to generate its representation $f(x_i; \theta)$. Then, it takes the dot product of the data representation and every class representation. This way, it calculates the similarity between the input data and classes. 
An activation function $\sigma$ is applied to the dot product to get the predicted probabilities of $x_i$:
\begin{equation}
    g(x_i;\theta, \omega) = \sigma([f(x_i; \theta) \circ \gamma(w_c;\omega)]_{w_c\in \mathcal{W}}).
\end{equation}
The choice of activation function is the same as defined in Section~\ref{sec:preliminary}.

\smallsection{Alternating Encoder Training} 
With the new model design, we rewrite the learning objective in Equation~\ref{eq:general-local-objective} as:
\begin{equation}
    {\mathcal{L}}_{m}(\theta, \omega) = \mathbb{E}_{(x_i, y_i) \sim \mathbf{D}_m}\ell[\sigma([f(x_i; \theta)\circ\gamma(w_c;\omega)]_{w_c\in \mathcal{W}}), y_i].
    \label{eq:loss-identified}
\end{equation}

The two encoders are two branches in the model. We want the representations obtained by one encoder to regulate the training of the other while preventing mutual interference. Therefore, at each local update step, we first fix the parameters in the label encoder and update the data encoder. Then, we fix the data encoder and update the label encoder. Let $\theta^{(t)}_{m,j}$ and $\omega^{(t)}_{m,j}$ be the parameters of the local data encoder and label encoder at $j$-th update step in $t$-th round and $\eta$ be the learning rate. The parameters are updated as:
\begin{equation}
    \theta^{(t)}_{m,j+1} \leftarrow \theta^{(t)}_{m,j} - \eta\nabla_{\theta^{(t)}_{m,j}}{\mathcal{L}}_{m}(\theta_{m,j}^{(t)},\omega_{m,j}^{(t)}),
\end{equation}
\begin{equation}
    \omega^{(t)}_{m,j+1} \leftarrow \omega^{(t)}_{m,j} - \eta\nabla_{\omega_{m,j}^{(t)}}{\mathcal{L}}_{m}(\theta^{(t)}_{m,j+1},\omega_{m,j}^{(t)}).
\end{equation}

\subsection{Anchor-Guided Alignment for Locally-Unaware Classes}
\label{sec:method-knowledge-distillation}

Due to the lack of label information of certain classes to support supervision, the training at each client is biased toward the identified classes~\cite{luo2021no,zhang2019category}. To mitigate such drift, we further exploit the global class representations to assist the alignment for locally-unaware classes. Since we formulate the classification problem as a matching between representations of classes and local data at each client, the class representations produced by the global label encoder can reflect the global distribution. Therefore, we regard the global class representations as anchors and use them to identify features for unaware classes at each client. Specifically, at the beginning of each round of local training, the client measures the distances between class anchors and local data representations. The nearest and farthest samples from the anchors are annotated. An additional loss term is added to the local optimization objective to reduce the distribution mismatch. Compared with common practices of pseudo-labeling~\cite{mclachlan1975iterative,lee2013pseudo} which assign labels based on model predictions, the annotations assigned by our anchor-guided method are independent of the biased classifier and are thus more reliable.

\begin{figure}[t]
    \centering
    \subfigure[Selecting positive samples based on global class anchors]{
	\centering
	\includegraphics[width=0.38\linewidth]{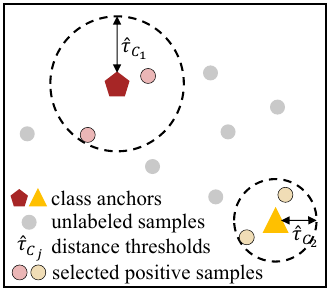}
	\label{fig:method-adaptation}
    }
    \subfigure[Effect of matching and alignment by using the global class representations as anchors]{
	\centering
	\includegraphics[width=0.57\linewidth]{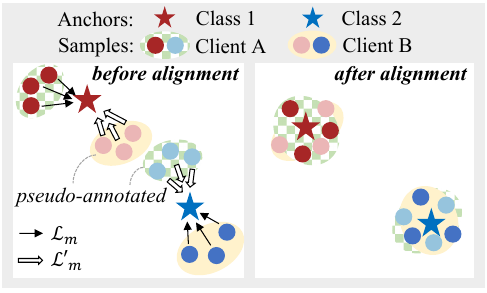}
	\label{fig:method-alignment-effect}
    }
    \caption{(a) illustrates how positive samples are annotated for locally-unaware classes based on distances to class anchors. (b) shows the effect of matching and alignment.}
    \label{fig:method-alignment}
\end{figure}

\smallsection{Deriving Class Anchors}
When the client receives the parameters of the label encoder $\omega^{(t)}$ at $t$-th round, it uses the latest label encoder to derive the global class anchors: $\{\gamma(w_c;\omega^{(t)}) \vert w_c \in \mathcal{W}\}$. 

\smallsection{Selecting Samples for Locally-Unaware Classes}
Client $m$ uses the received data encoder to generate representations of its local data: $\{f(x_i;\theta^{(t)}) \vert x_i \in \mathcal{X}_m\}$.
Then, the client calculates the cosine distance from every class anchor to the local data in latent space:
\begin{equation}
    d_{i,c}^{(t)} = 1 - \frac{\gamma(w_{c};\omega^{(t)}) \circ f(x_i; \theta^{(t)})}{\Vert \gamma(w_{c};\omega^{(t)}) \Vert_2 \cdot \Vert f(x_i;\theta^{(t)}) \Vert_2 }.
\end{equation}

Then, the client annotates samples for the locally-unaware classes $\overline{\mathbf{C}_m}$ based on the distances. Samples with the closest distances to the class anchor $\gamma(w_{c};\omega^{(t)})$ are annotated as positive samples of class $c$. Similarly, samples that are farthest from $\gamma(w_{c};\omega^{(t)})$ are annotated as negative samples of $c$. The number of samples to be annotated depends on the percentile of distances. We define two thresholds, $\hat{\tau}_{c}^{(t)}$ and $\check{\tau}_{c}^{(t)}$, as the $q_1$-th and $q_2$-th percentile of the distances over all samples for annotating positive and negative samples respectively. The client annotates the samples whose distances are less than $\hat{\tau}_{c}^{(t)}$ as positive samples (i.e., $\tilde{y}^{(t)}_{i,c} = 1$) and those with distances greater than $\check{\tau}_{c}^{(t)}$ as negative samples (i.e., $\tilde{y}^{(t)}_{i,c} = 0$). Figure~\ref{fig:method-adaptation} shows an example of selecting positive samples for two classes. The dataset for alignment after the $t$-th round is as follows:
\begin{equation}
    {\mathbf{D}'}_{m}^{(t)} \leftarrow \{(x_i, \tilde{y}_{i}^{(t)})\vert d^{(t)}_{i,c} < \hat{\tau}_{c}^{(t)} \text{ or } d^{(t)}_{i,c} > \check{\tau}_{c}^{(t)}, c \in \overline{\mathbf{C}_m}\}.
\end{equation}

For single-label classification, we add another constraint that a sample whose true label is not in $\mathbf{C}_m$ is annotated as a positive sample of class $c\in \overline{\mathbf{C}_m}$ only if $c$ is the closest to it among all classes.

\smallsection{Alignment}
The annotations for unaware classes are then used to guide the alignment at client $m$. We add an additional loss term to the local learning objective. The loss over ${\mathbf{D}'}_m^{(t)}$ is as follows:
\begin{equation}
\resizebox{.9\linewidth}{!}{$
    {\mathcal{L}'}_{m}^{(t)}(\theta, \omega) = \mathbb{E}_{(x_i, \tilde{y}_i) \sim {\mathbf{D}'}_m^{(t)}}\ell'[\sigma([f(x_i; \theta)\circ\gamma(w_c;\omega)]_{w_c\in \mathcal{W}}, \tilde{y}_i],
$}
\label{eq:loss-unaware}
\end{equation}
where $\ell'$ represents the loss function with the same choice as defined in Equation~\ref{eq:cross-entropy-loss} and \ref{eq:binary-cross-entropy-loss}. A slight difference is that $\ell'$ here is summed over $\overline{\mathbf{C}_m}$. Finally, the local learning objective is to jointly minimize Equation~\ref{eq:loss-identified} and \ref{eq:loss-unaware}, i.e., $\text{min}_{\theta, \omega}[\mathcal{L}_{m}(\theta, \omega) + {\mathcal{L}'}_{m}^{(t)}(\theta, \omega)]$. Figure~\ref{fig:method-alignment-effect} illustrates the effect of these two losses. 
\section{Analysis on Generalization Bound}
In this section, we perform an analysis of the generalization performance of the aggregated model in federated learning. 

Denote $\mathcal{D}$ as the global distribution on input space $\mathcal{X}$, and $\tilde{\mathcal{D}}$ as the induced global distribution over feature space $\mathcal{Z}$. Similarly, for the $m$-th client, denote $\mathcal{D}_m$ as the local distribution and $\tilde{\mathcal{D}}_m$ be the induced image of $\mathcal{D}_m$ over $\mathcal{Z}$. We review a typical theoretical upper bound for the generalization of global hypothesis~\cite{peng2019federated,zhu2021data,lin2020ensemble}:

\begin{theorem}[Generalization Bound of Federated Learning]
\label{theorem:generalization-FL}
Assume there are $M$ clients in a federated learning system. Let $\mathbf{H}$ be the hypothesis class with VC-dimension $d$. The global hypothesis is the aggregation of $h_m$, i.e., $h = \frac{1}{M}\sum_{m\in [M]}h_m$. Let $\mathcal{L}(h)$ denote the expected risk of $h$. With probability at least $1-\delta$, for $\forall h \in \mathbf{H}$:
\begin{equation}
\resizebox{.9\linewidth}{!}{$
    \begin{aligned}
    \mathcal{L}(h) \leq & \frac{1}{M}\sum\nolimits_{m\in [M]}\hat{\mathcal{L}}_m(h_m) + \frac{1}{M}\sum\nolimits_{m\in [M]}[d_{\mathbf{H}\Delta \mathbf{H}}(\tilde{\mathcal{D}}_m, \tilde{\mathcal{D}})+\lambda_m]\\
    &+ \sqrt{\frac{4}{N}(d\mathop{log}\frac{2eN}{d} + \mathop{log}\frac{4M}{\delta})},
    \end{aligned}
$}
\end{equation}
where $\hat{\mathcal{L}}_m(h_m)$ is the empirical risk on the $m$-th client given $N$ observed samples, $d_{\mathbf{H}\Delta \mathbf{H}}(\cdot, \cdot)$ is the $\mathcal{A}$-distance that measures the divergence between two distributions based on the symmetric difference with respect to $\mathbf{H}$, $\lambda_m$ is the risk of the optimal hypothesis over $\mathbf{H}$ with respect to $\mathcal{D}$ and $\mathcal{D}_m$, $e$ is the base of the natural logarithm.
\end{theorem}

Theorem~\ref{theorem:generalization-FL} applies to the traditional algorithm FedAvg~\cite{mcmahan2017communication}, we observe two factors that affect the quality of the global hypothesis: the divergence between the local and global distributions $d_{\mathbf{H}\Delta \mathbf{H}}(\tilde{\mathcal{D}}_m, \tilde{\mathcal{D}})$ and the sample size $N$. Then, we discuss the generalization bound when \our introduces empirical distributions for locally-unaware classes to align latent spaces.

\begin{corollary}[Generalization Bound of Federated Learning with Mix-up Distributions]
Let $\mathcal{D}'_m$ denote the distribution added for aligning the $m$-th client. Define the mix-up distribution $\mathcal{D}^*_m$ to be a mixture of the original local distribution $\mathcal{D}_m$ and $\mathcal{D}'_m$:
\begin{equation}
    \mathcal{D}^*_m = \alpha\mathcal{D}_m + (1-\alpha)\mathcal{D}'_m,
\end{equation}
where $\alpha \in [0, 1]$ is the weight of the original distribution, which is decided by the number of empirical samples added. Let $\mathbf{H}$ be the hypothesis class with VC-dimension $d$. The global hypothesis is the aggregation of $h_m$, i.e., $h = \frac{1}{M}\sum_{m\in [M]}h_m$. With probability at least $1-\delta$, for $\forall h \in \mathbf{H}$:
\begin{equation}
\resizebox{.9\linewidth}{!}{$
    \begin{aligned}
    \mathcal{L}(h) \leq & \frac{1}{M}\sum\nolimits_{m\in [M]}\hat{\mathcal{L}}_m(h_m)\\
    &+ \frac{1}{M}\sum\nolimits_{m\in[M]}[\alpha d_{\mathbf{H}\Delta \mathbf{H}}(\tilde{\mathcal{D}}_m, \tilde{\mathcal{D}}) + (1-\alpha)d_{\mathbf{H}\Delta \mathbf{H}}(\tilde{\mathcal{D}}'_m, \tilde{\mathcal{D}}) + \lambda_m]\\
    &+ \sqrt{\frac{4}{N^*}(d\mathop{log}\frac{2eN^*}{d} + \mathop{log}\frac{4M}{\delta})},
    \end{aligned}
$}
\end{equation}
where $\hat{\mathcal{L}}_m(h_m)$ is the empirical risk on the $m$-th client given $N^*$ ($N^* > N$) observed samples, $e$ is the base of the natural logarithm. 

\end{corollary}

By combining the local dataset with pseudo-annotated samples, \our increases the sample size i.e., $N^* > N$, thus the last term of the bound becomes smaller. Second, given that the selected samples are in proximity to the anchors which are derived by the ensemble of the empirical distributions across all clients, the distribution derived via class anchors would exhibit lower divergence from the global distribution compared to the original local distribution i.e., $d_{\mathbf{H}\Delta \mathbf{H}}(\tilde{\mathcal{D}}'_m, \tilde{\mathcal{D}}) < d_{\mathbf{H}\Delta \mathbf{H}}(\tilde{\mathcal{D}}_m, \tilde{\mathcal{D}})$. The proof and more details are given in Appendix. Therefore, \our can achieve a better generalization bound than traditional methods~\cite{mcmahan2017communication}, suggesting a strong potential for performance improvement.
\section{Experiments}\label{sec:experiment}
\subsection{Datasets}

We conduct experiments on 6 datasets covering 4 different application scenarios and both single-label and multi-label classification problems. Table~\ref{tab:dataset} offers an overview and the details are as follows. 
\begin{enumerate}[nosep,leftmargin=*]
    \item \textbf{Behavioral Context Recognition.} The task is to infer the context of human activity. \textbf{ExtraSensory}~\cite{vaizman2017recognizing} is a benchmark dataset for this task. The classes can be partitioned into 5 categories (e.g. location, activity, etc.). Based on ExtraSensory, we construct 3 datasets with non-overlapping client class sets.
        \textbf{ES-5}: 
            We set 5 clients and every client only has annotations from a different category (i.e., one category to one client). Training samples are then assigned to clients according to their associated classes. Since ExtraSensory is a multi-label dataset, we assign samples based on the most infrequent class among multiple labels to ensure each locally-identified class will have at least one positive sample. To make this dataset more realistic, we always assign all data of a subject to the same client.
        \textbf{ES-15 and ES-25}:
            We increase the number of clients to 15 and 25 to further challenge the compared methods. We start with the 5 class groups as ES-5 and iteratively split the groups until the number of class groups matches the number of clients. During every split, we select the group with the most classes and randomly divide it into two sub-groups. 
            Every class group is visible and only visible to one client.
            One can then apply a similar process as ES-5 to assign training samples to clients. 
    \item \textbf{Medical Code Prediction.} Medical codes describe whether a patient has a specific medical condition or is at risk of development. The task is to annotate medical codes from clinical notes. We start with the MIMIC-III database~\cite{johnson2016mimic} and follow the preprocessing method in \cite{mullenbach2018explainable} to form the benchmark MIMIC-III 50-label dataset. The classes span 10 categories in the ICD-9 taxonomy\footnote{the International Statistical Classification of Diseases and Related Health Problems (ICD): \url{ftp://ftp.cdc.gov/pub/Health_Statistics/NCHS/Publications/ICD-9/ucod.txt}}. We construct \textbf{MIMIC-III-10} by partitioning the dataset into 10 clients following the same strategy as in ES-5.
    \item \textbf{Human Activity Recognition.} The task aims at identifying the movement or action of a person based on sensor data.
    We start with the PAMAP2~\cite{reiss2012introducing} dataset, which collects data of physical activities from 9 subjects. We construct the \textbf{PAMAP2-9} dataset by regarding each subject as a client. For each client, we randomly select 5 classes to be its locally-identified classes. 
    \item \textbf{Text Classification.} We use the Reuters-21578 R8 dataset~\cite{2007:phd-Ana-Cardoso-Cachopo}, which consists of news articles classified into 8 categories. We construct \textbf{R8-8} by randomly partitioning the data into 8 subsets and assigning one subset to each client. 
    For each client, we randomly select 3 classes to be the identified classes.
\end{enumerate}

More details about data preprocessing are described in Appendix.

\subsection{Compared Methods}
We compare \our with classical~\cite{mcmahan2017communication} and state-of-the-art federated learning methods for non-IID data~\cite{li2020federated, li2021model,karimireddy2020scaffold} as follows.
\begin{itemize}[leftmargin=*]
    \item \textbf{FedAvg}~\cite{mcmahan2017communication} is a classical federated learning algorithm where the server averages the updated local model parameters in each round to obtain the global model.
    \item \textbf{FedProx}~\cite{li2020federated} enforces a $L_2$ regularization term in local optimization which limits the distance between global and local models.
    \item \textbf{MOON}~\cite{li2021model} adds a contrastive loss term to maximize the consistency of representations learned by the global and local models and minimize the consistency between representations learned by the local models of consecutive rounds.
    \item \textbf{Scaffold}~\cite{karimireddy2020scaffold} maintains control variates to estimate the update directions of global and local models. The drift in local training is approximated by the difference between the update directions. This difference is then added to the local updates to mitigate drift.
\end{itemize}

We also compare two state-of-the-art methods designed for federated classification with missing classes. Specifically,
\begin{itemize}[leftmargin=*, nosep]
    \item \textbf{FedRS}~\cite{li2021fedrs} is designed for the missing class scenario where each client owns data for part of the classes (i.e., \textit{locally-identified classes} in our terminology). It restricts the weight update of missing classes by adding scaling factors to the softmax operation.
    \item \textbf{FedPU}~\cite{lin2022federated} addresses scenarios where clients annotate only a small portion of their data, and unlabeled data exists for both locally identified and unaware classes. 
    It leverages the labeled data at every client to estimate the misclassification loss of unaware classes from other clients and incorporates this estimated loss into local optimization objectives. 
\end{itemize}

\subsection{Experimental Setup}
\smallsection{Base Neural Network Model}
For a fair comparison, we use the same model setting for all compared methods. 
The data encoder is based on the Transformer architecture~\cite{vaswani2017attention} with one encoder layer. 
There are 4 attention heads, and the dimension of the feed-forward network is 64. 
The label encoder is a single hidden layer neural network. 
The dimension $d$ of representations is 256. 
Since the size of the label encoder is equivalent to the classifier layer in the conventional classification model, there is no extra overhead during model communication in \our. Additionally, when considering future work involving the use of advanced neural language models as the label encoder, we can train only the adapter module~\cite{houlsby2019parameter}, i.e., adding a small number of new parameters to the pretrained language model. The adapters are transferred and aggregated, while the other layers remain fixed at all parties.

\begin{table}[t]
\centering
    \caption{Dataset statistics. The imbalance factor refers to the ratio of the smallest class size to the largest class size.}
\resizebox{\linewidth}{!}{
    \begin{tabular}{rcccccc}
    \toprule
    \textbf{Dataset} & $|C|$ & \textbf{\# of clients} & \textbf{Avg.} $N_m$ & \textbf{Avg.} $|C_m|$ & \textbf{Imbalance} & \textbf{Remarks}\\
    \midrule
    \textbf{ES-5}  & \multirow{3}[4]{*}{51} & 5 & 5,446 & 10.2 & \multirow{3}[4]{*}{0.0013} & \multirow{4}[4]{1.9cm}{Multi-label, non-overlapping client class sets}\\
    \cmidrule{1-1} \cmidrule{3-5}
    \textbf{ES-15} & & 15 & 1,769 & 3.4 & \\
    \cmidrule{1-1} \cmidrule{3-5}
    \textbf{ES-25} & & 25 & 1,073 & 2.04 & \\
    \cmidrule{1-6}
    \textbf{MIMIC-III-10} & 50 & 10 & 807 & 5 & 0.1157\\
    \midrule
    \textbf{PAMAP2-9} & 18 & 9 & 1,287 & 5 & 0.2049 & \multirow{2}[2]{1.9cm}{Single-label} \\
    \cmidrule{1-6}
    \textbf{R8-8} & 8 & 8 & 617 & 3 & 0.0130 &\\
    \bottomrule
    \end{tabular}
}
    \label{tab:dataset}
\end{table}

\begin{table*}[t]
  \centering
  \caption{Main experimental results (\% averaged over 5 runs). ES-5, ES-15, ES-25 and MIMIC-III-10 are multi-label datasets where class sets across clients have no overlap. PAMAP2-9 and R8-8 are single-label datasets where client class sets overlap. 
  }
\scalebox{.94}{
    \begin{tabular}{lcccccccccccc}
    \toprule
    \multirow{2}[4]{*}{Method} & \multicolumn{2}{c}{ES-5} & \multicolumn{2}{c}{ES-15} & \multicolumn{2}{c}{ES-25} & \multicolumn{2}{c}{MIMIC-III-10} & \multicolumn{2}{c}{PAMAP2-9} & \multicolumn{2}{c}{R8-8}\\
    \cmidrule(lr){2-3} \cmidrule(lr){4-5} \cmidrule(lr){6-7} \cmidrule(lr){8-9} \cmidrule(lr){10-11} \cmidrule(lr){12-13} & F1 & Acc & F1 & Acc & F1 & Acc & F1 & Acc & F1 & Acc & F1 & Acc \\
    \midrule
    FedAvg~\cite{mcmahan2017communication} & 28.77 & 80.79 & 22.71 & 62.44 & 19.52 & 50.42 & 35.04 & 67.07 & 68.89 & 71.45 & 78.51 & 92.76\\
    FedProx ($\mu=0.001$)~\cite{li2020federated} & 29.26 & 80.67 & 22.42 & 61.91 & 19.48 & 51.16 & 30.57 & 61.76 & 69.70 & 73.63 & 79.05 & 92.61\\
    FedProx ($\mu=0.0001$)~\cite{li2020federated} & 28.53 & 79.14 & 22.52 & 61.95 &   19.05 & 52.33 & 33.73 & 64.31 & 69.38 & 71.78 & 75.98 & 91.78\\
    MOON~\cite{li2021model} & 29.12 & 81.00 & 22.84 & 62.53 & 19.52 & 50.24 & 34.62 & 66.34 & 71.70 & 74.25 & 79.26 & 93.07\\
    Scaffold~\cite{karimireddy2020scaffold} & 28.14 & 77.13 & 23.15 & 61.69 & 19.73 & 48.81 & 33.58 & 62.84 & 73.57 & 75.60 & 82.83 & 94.43\\
    \midrule
    FedPU~\cite{lin2022federated} & 27.95 & 79.82 & 22.27 & 59.22 & 16.97 & 34.59 & 33.23 & 62.83 & 85.50 & 87.66 & 83.06 & 94.17\\
    FedRS ($\alpha=0.5$)~\cite{li2021fedrs} & 28.01 & 78.72 & 22.50 & 62.09 & 19.44 &	51.41 & 34.82 & 66.80 & 68.70 & 71.42 & 80.10 & 92.74\\
    FedRS ($\alpha=0.9$)~\cite{li2021fedrs} & 28.25 & 79.25 & 22.55 & 62.17 & 19.40 & 50.87 & 35.44 & 67.45 & 71.81 & 74.44 & 76.68 & 91.81\\
    \midrule
    \our & \textbf{30.19} & \textbf{84.05} & \textbf{23.36} & \textbf{73.61} & \textbf{20.80} & \textbf{67.12} & \textbf{37.97} & \textbf{74.37} & \textbf{87.21} & \textbf{88.14} & \textbf{83.76} & \textbf{94.92} \\
    \bottomrule
    \end{tabular}
}
\label{tab:exp-main-result}
\end{table*}

\begin{figure*}[t]
    \centering
    \includegraphics[width=\linewidth]{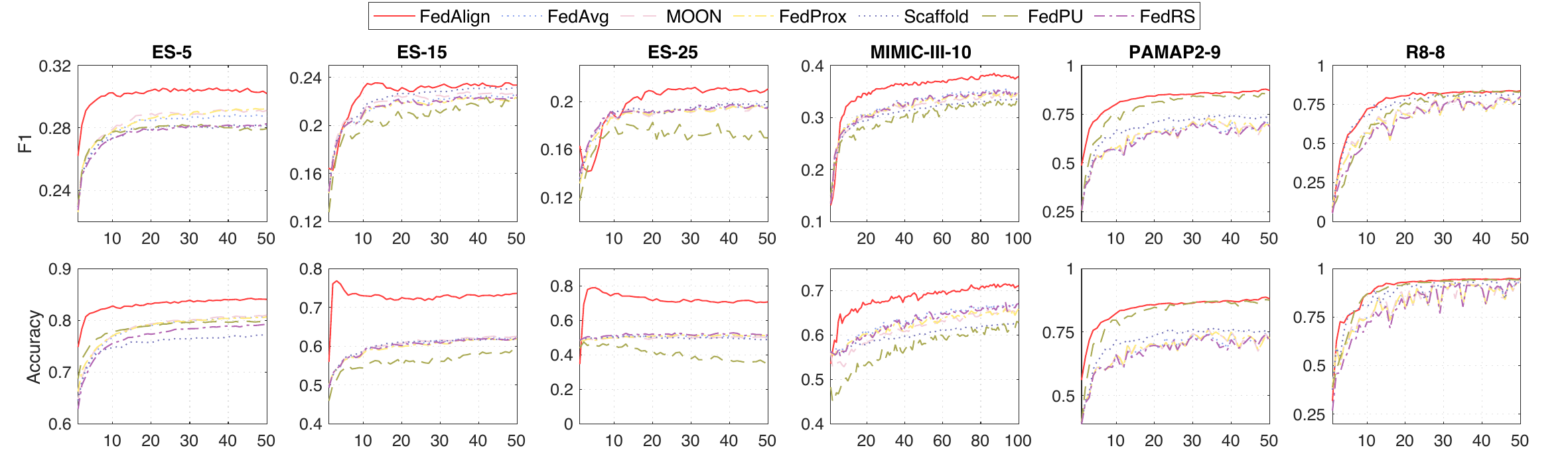}
    \caption{Performance w.r.t. communication rounds on six datasets. The results are averaged over 5 runs.}
\label{fig:exp-communication-round}
\end{figure*}

\smallsection{Evaluation Metrics}
Due to label imbalance, we adopt both accuracy and F1-score to evaluate the performance. They are often used as benchmark metrics for the datasets and tasks in our experiments~\cite{vaizman2017recognizing,gennatas2020expert,reiss2012introducing,piao2022sparse}. We calculate the metrics for each class and report the macro-average. All experiments are repeated 5 times with a fixed set of random seeds for all compared methods. 

\smallsection{Train/Test Split}
We set aside a portion of the dataset for testing the global model. For MIMIC-III and R8, we use the data split provided by the dataset. For the other datasets, we use 20\% of the data for testing and distribute the rest of the data to clients for training.

\smallsection{Federated Learning Setting}
For ES-5, ES-15, ES-25, PAMAP2-9 and R8-8, we run $T=50$ rounds. For MIMIC-III-10, we run $T=100$ rounds as it takes longer to converge. The number of selected clients per round is $\lvert S_t\rvert = 5$ and the local epochs $E=5$. Note that we conduct sensitivity analysis in Section~\ref{sec:exp-sensitivity} and show the conclusion of the results is robust to the value of $\lvert S_t\rvert$ and $E$.

\smallsection{Hyperparameters}
For the compared methods, we try different values for the hyperparameters $\mu$ in FedProx and MOON, and $\alpha$ in FedRS, that are often adopted in the previous papers~\cite{li2020federated,li2021model,li2021fedrs}. The values are displayed alongside the method name in Table~\ref{tab:exp-main-result}.

\subsection{Main Results and Analysis}
\smallsection{Multi-Label, Non-overlapping Client Class Sets}
Table~\ref{tab:exp-main-result} shows the results. As one can clearly see, \our always yields better performance than the baseline methods. Remarkably, with non-identical client class sets, the three state-of-the-art algorithms designed to deal with non-IID data (i.e., FedProx, MOON, and Scaffold) do not guarantee improvement over FedAvg (e.g., Scaffold loses to FedAvg on ES-5). In addition, although FedRS and FedPU are designed for missing class scenarios, their mechanisms are specifically tailored for single-label classification. In the context of multi-label classification, the label of one class does not indicate the labels of other classes, and the weight update of a class is solely influenced by its own features. Therefore, the scaling factors in FedRS and the misclassification loss estimation in FedPU become ineffective. 

\smallsection{Single-Label, Non-identical but Overlapping Client Class Sets}
\our outperforms the baselines on both applications. The non-IID problems that FedRS and FedPU aim to tackle (i.e., missing class scenario, and positive and unlabeled data) are slightly different from ours. Although they show improvements over FedAvg and methods designed for the typical non-IID setting (i.e., FedProx, MOON, and Scaffold), \our shows better performance compared with FedRS and FedPU in the problem of non-identical client class sets.

\smallsection{Performance w.r.t. Communication Rounds}
Figure~\ref{fig:exp-communication-round} shows the test performance with respect to communication rounds. 
\our shows its advantage from the early stages of training. 
This indicates the pretrained text representations provide good initialization for the label encoder to guide the alignment of latent spaces. We do notice a decrease in the F1-score of \our on ES-25 during initial rounds. This can be attributed to the noise in pseudo annotations for locally-unaware classes due to the undertrained encoders. However, as the training progresses, the quality of the pseudo annotations improves, leading to enhanced performance.

\begin{table}[t]
  \centering
  \caption{F1-Score (\% Averaged Over 5 Runs) of Ablation Study}
    \begin{tabular}{lcccc}
    \toprule
    Method & ES-25 & MIMIC-III-10 & PAMAP2-9 & R8-8\\
    \midrule
    FedAvg & 19.52 & 35.04 & 68.89 & 78.51\\
    \our w/o AL & 19.57 & 37.87 & 70.87 & 79.67\\
    \our w/o SE & 19.62 & 34.91 & 83.39 & 82.37 \\
    \our w/o AT & 20.28 & 37.09 & 86.01 & 83.22 \\
    \our & \textbf{20.80} & \textbf{37.97} & \textbf{87.21} & \textbf{83.76}\\
    \bottomrule
    \end{tabular}
  \label{tab:exp-ablation-study}
\end{table}

\subsection{Ablation Study}
We conduct an ablation study to evaluate the contribution of each design in \our. First, we evaluate the performance of the method without alignment for locally-unaware classes (denoted as \textbf{\ourbold w/o AL}). The classification model consists of a data encoder and a label encoder and the framework conducts alternating training of the two modules. Second, we evaluate the performance of the method without the semantic label name sharing (denoted as \textbf{\ourbold w/o SE}). In this case, the dataset for alignment is formed by annotating the samples according to prediction confidence given by the latest global model. For locally-unaware classes, samples with high prediction confidence are pseudo-annotated, and the confidence thresholds are decided by the same percentile values as in \our. Third, we evaluate the performance of the method without alternating training (denoted as \textbf{\ourbold w/o AT}) which updates label and data encoders simultaneously.

Since the model aggregation method in \our is based on FedAvg (i.e., averaging the model parameters), we also compare FedAvg as the baseline method. Table~\ref{tab:exp-ablation-study} shows the F1-scores. We notice the performance decreases when removing any of the designs. This suggests the designs in \our all contribute to improvement, and combining them can produce the best performance.

\begin{figure}[t]
    \centering
    \subfigure[Performance w.r.t. Participating Clients Per Round]{
    \centering
    \includegraphics[width=\linewidth]{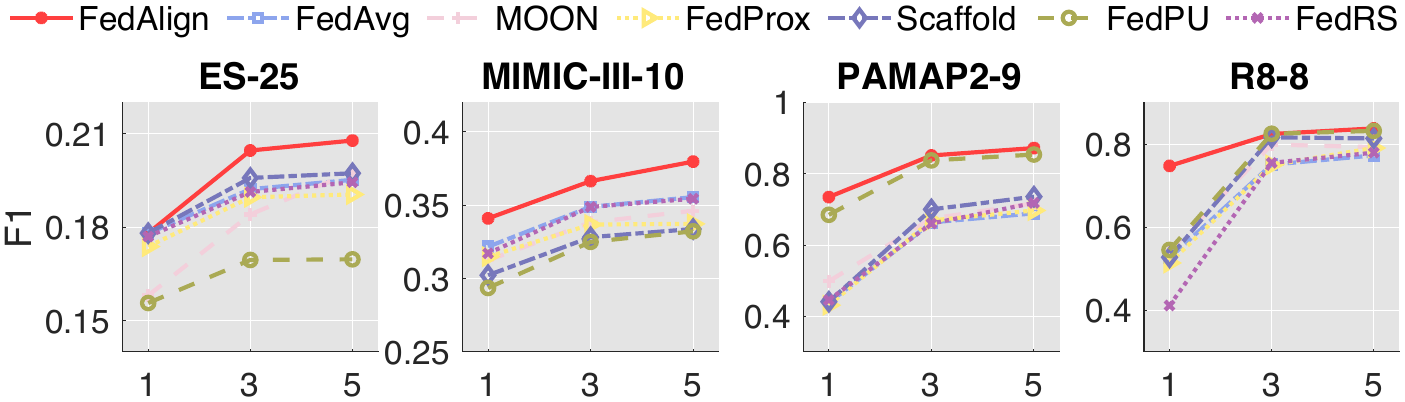}
    \label{fig:exp-nclients}
    }
    \subfigure[Performance w.r.t. Local Training Epochs]{
    \centering
    \includegraphics[width=\linewidth]{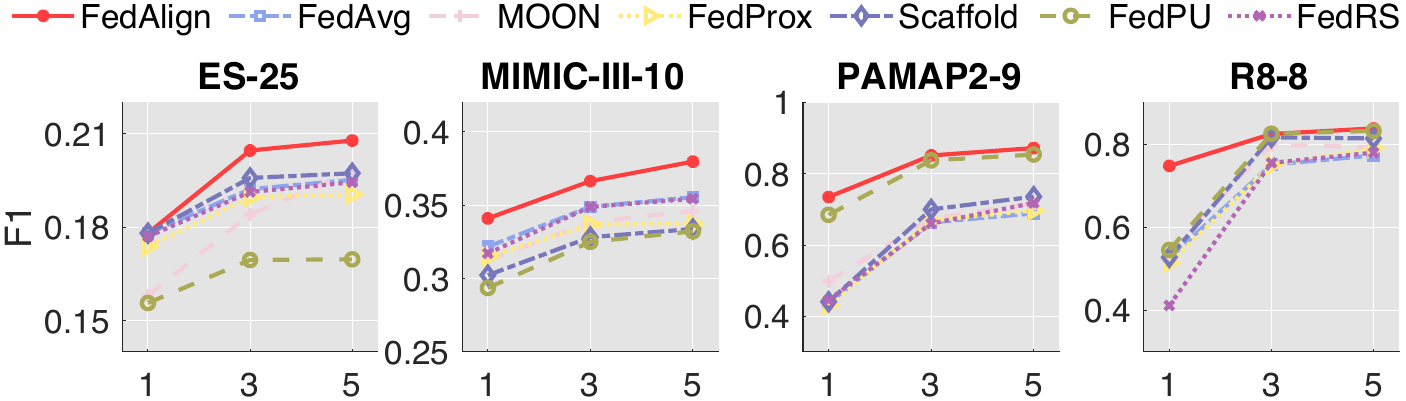}
    \label{fig:exp-nepochs}
    }
    \subfigure[Performance w.r.t. Distance Threshold]{
    \centering
    \includegraphics[width=\linewidth]{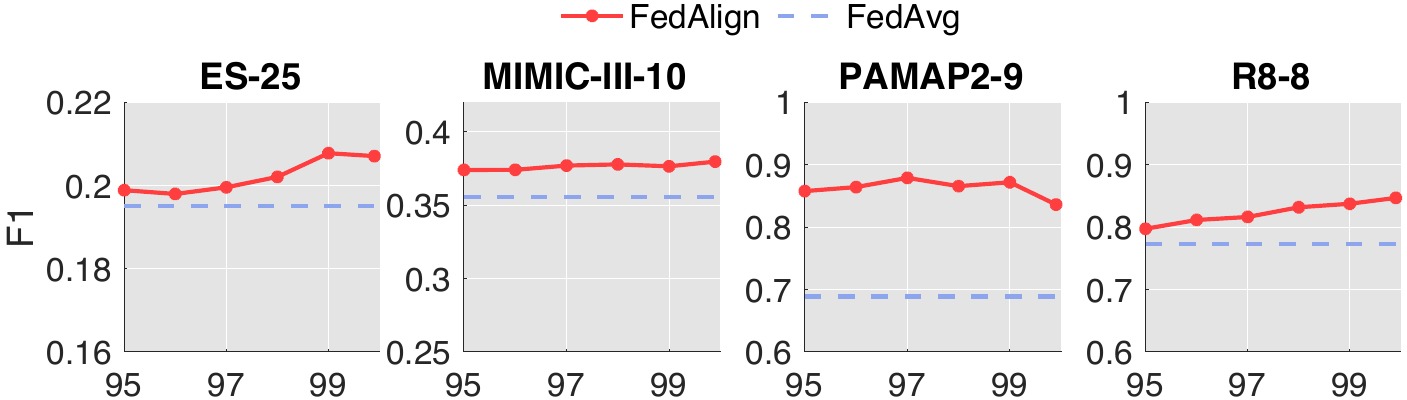}
    \label{fig:exp-distance}
    }
    \caption{Sensitivity Analysis}
\end{figure}

\subsection{Sensitivity Analysis}
\label{sec:exp-sensitivity}

\smallsection{Participating Clients Per Round}
The number of participating clients in each round (i.e., $\lvert S_t\rvert$) has an effect on the speed of convergence~\cite{li2019convergence}. We vary $\lvert S_t\rvert$ from 1 to 5 and compare \our with all baseline methods. The comparisons in F1-score are shown in Figure~\ref{fig:exp-nclients}. We observe that \our can always outperform the baseline methods under different values of $\lvert S_t\rvert$. 

\smallsection{Local Training Epochs}
We vary the local training epochs from 1 to 5 and compare the performance of \our with all baseline methods. The comparisons are shown in Figure~\ref{fig:exp-nepochs}. We see that \our has consistently better performance than the baselines.

\smallsection{Distance Threshold for Selecting Samples for Unaware Classes}
In Section~\ref{sec:method-knowledge-distillation}, we set the threshold for assigning labels to samples for locally-unaware classes based on distance percentiles. To test the robustness of \our to this hyperparameter, we vary the threshold for annotating positive samples by using different percentiles (95 to 99.9). Figure~\ref{fig:exp-distance} shows the result. We see that \our only needs a very small amount of pseudo annotations to have significant improvements over FedAvg. Notably, samples closer to the class anchors exhibit a higher likelihood of being accurately annotated, providing better guidance for alignment.

\subsection{Case Studies}

\begin{figure}[t]
    \centering
    \includegraphics[width=1.05\linewidth]{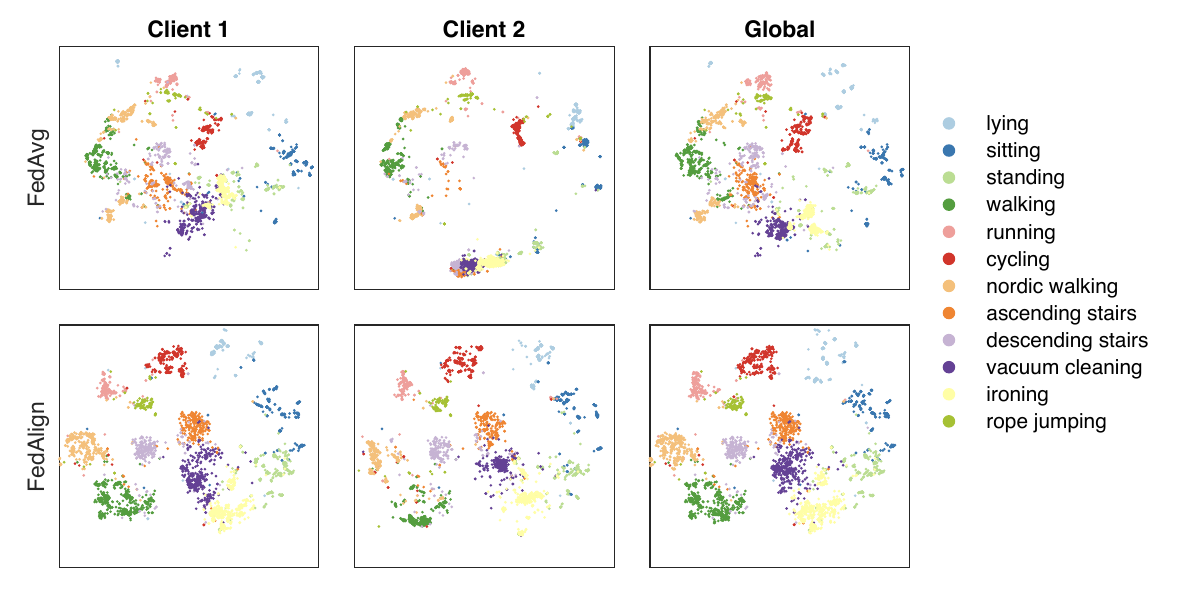}
    \caption{Data representations generated by two local models and the global model on the testing set of PAMAP2-9.}
\label{fig:exp-latent-space}
\end{figure}

\begin{figure}[t]
    \centering
    \subfigure[Similarity of Class Representations]{
    \centering
    \includegraphics[width=0.47\linewidth]{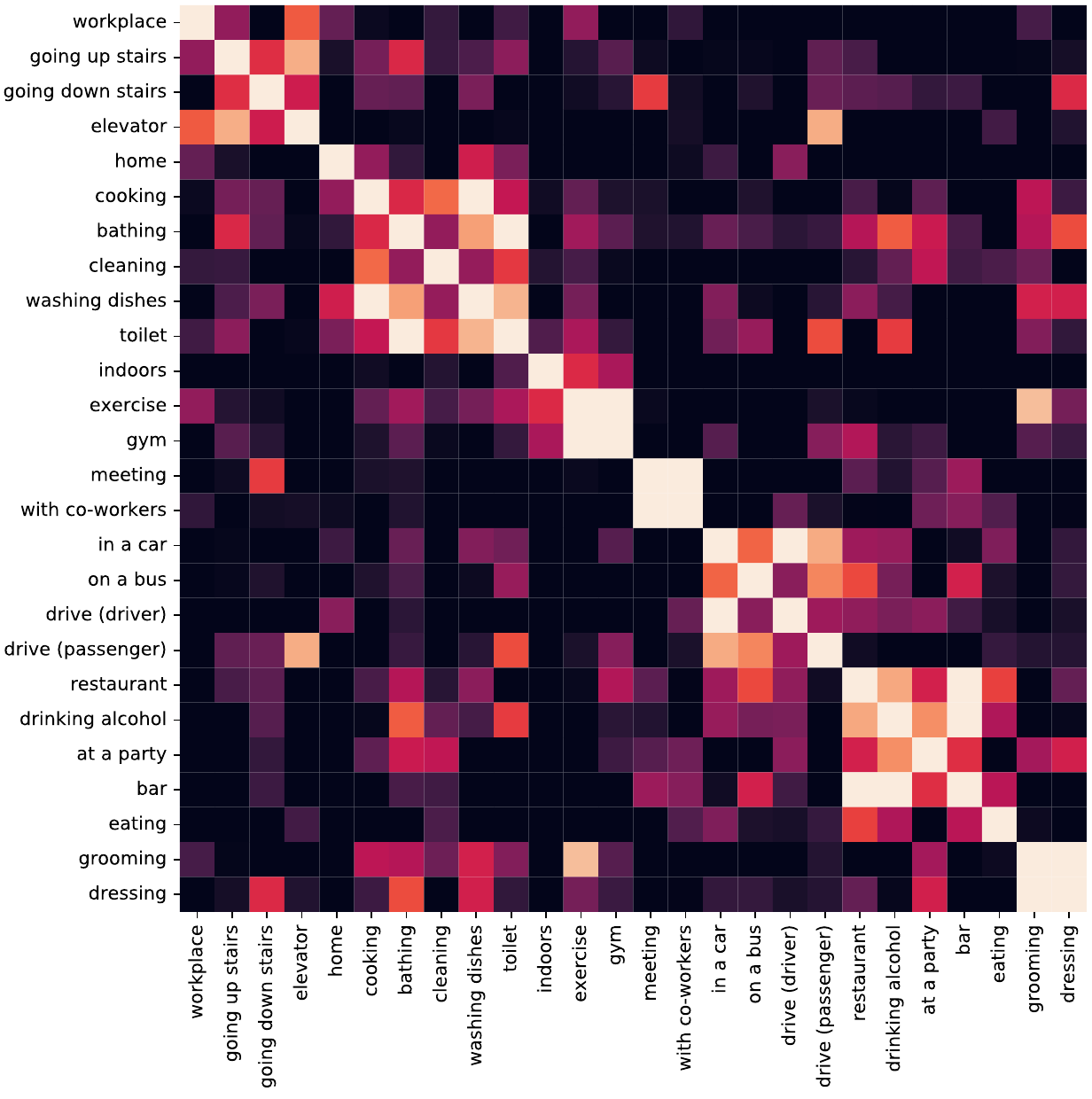}
    \label{fig:exp-class-rep}
    }
    \subfigure[PMI of Labels in Centralized Dataset]{
    \centering
    \includegraphics[width=0.47\linewidth]{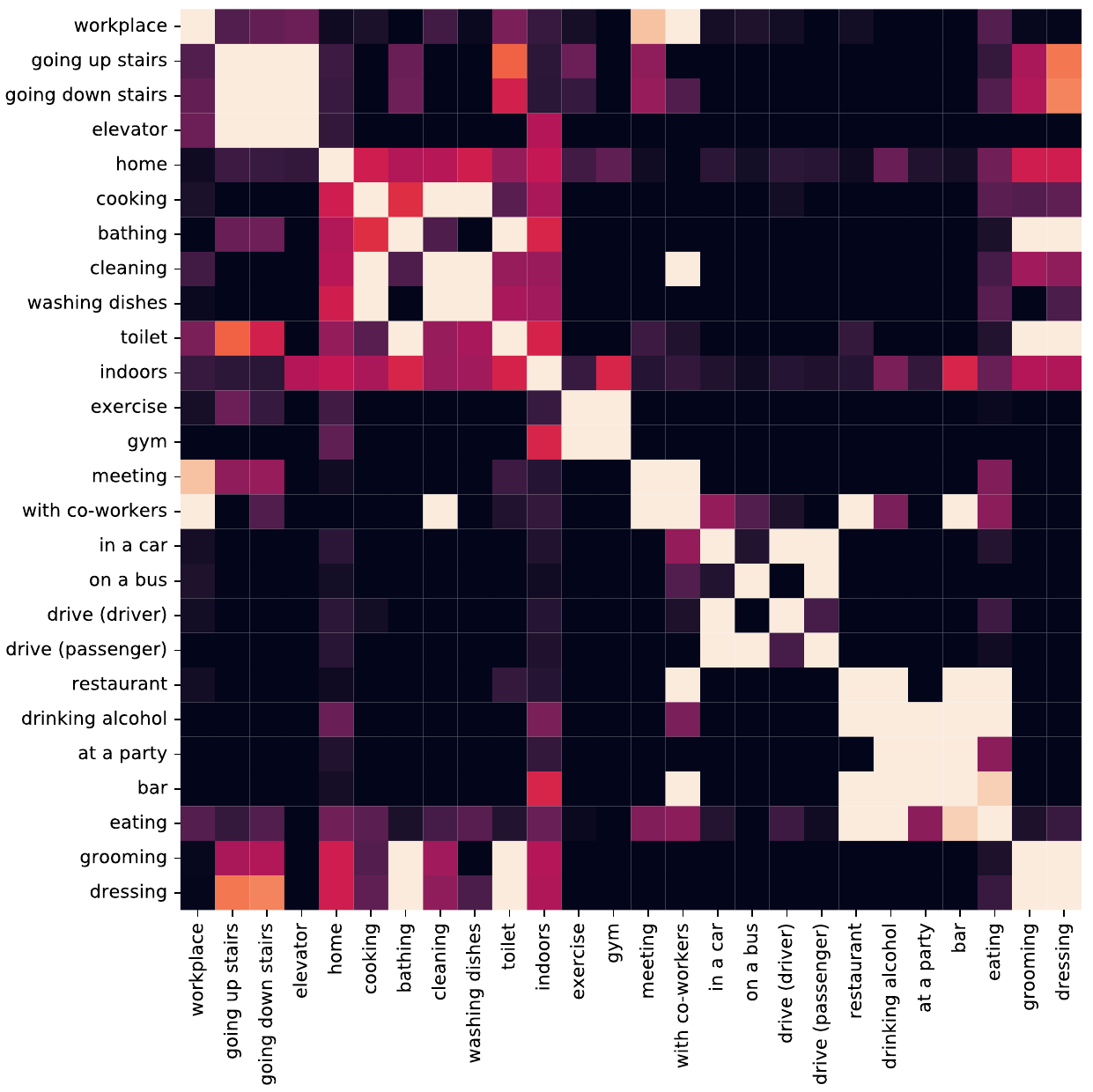}
    \label{fig:exp-label-cooccur}
    }
    \caption{(a) shows cosine similarities among class representations of ES-25 learned via \our. (b) demonstrates the PMI of labels in the centralized dataset as a reference of ground truth. Brighter colors indicate higher similarity/PMI.
    }
\end{figure}

\smallsection{Visualization of Feature Latent Spaces}
We visualize the learned data representations in PAMAP2-9. We generate the data representations on the testing set by the global model and the local models of two participating clients after 50 communication rounds. The locally-identified classes at the two clients are \{walking, running, cycling, ironing, rope jumping\} and \{walking, lying, sitting, standing, vacuum cleaning\} respectively. There are one overlapping class and four client-exclusive classes per client. We use t-SNE~\cite{van2008visualizing} to project the representations to 2-dimensional embeddings and compare the learned representations by FedAvg and \our. In order to see if the representations generated by different client models are aligned by classes, for each algorithm, we gather the data representations generated by the client models and the global model together to perform the t-SNE transformation.
The visualization is shown in Figure~\ref{fig:exp-latent-space}. We position them in the same coordinates. When training via FedAvg, we observe that the data representations of the same class generated by the two local models are likely to fall into different locations in the latent space. This suggests that the latent spaces of the two clients are misaligned, leading to less discriminability among data representations from different classes in the global latent space after model aggregation. On the contrary, when training via \our, the data representations of the same class generated by the two local models have similar locations in latent space. In addition, the data representations learned by \our have clearer separations than those learned by FedAvg. 

\smallsection{Similarity Among Class Representations}
We then analyze the similarities among the class representations of ES-25 learned via \our. Recall that ES-25 is the multi-label classification task where the class sets at different clients are non-overlapping. We use the label encoder from the global model trained after 50 rounds to generate class representations. For a clear view of group similarities, we apply Spectral Clustering~\cite{shi2003multiclass} to cluster the class representations and sort the label names based on the assigned clusters. We visualize the cosine similarities of a subset of the classes as shown in Figure~\ref{fig:exp-class-rep}, where brighter colors indicate higher similarity. The observed similarity patterns in the class representations conform with our knowledge about what contexts of daily activities often happen together or not. For example, the representations of the classes, ``toilet'' and ``bathing'', ``meeting'' and ``with co-workers'', ``gym'' and ``exercise'' have higher similarity, while they have less similarity with other classes. 
To provide a reference for ground truth, we calculate the PMI of labels based on their co-occurrence in the centralized dataset to indicate how strong the association is between every two classes. We show the results in Figure~\ref{fig:exp-label-cooccur}. The brighter the color, the higher the PMI (i.e., the two classes have a stronger association). The order of the classes is the same as in Figure~\ref{fig:exp-class-rep}. We observe the two figures display similar patterns of associations among classes. Although the class sets of different clients are non-overlapping, the label encoder trained via \our successfully captures associations among classes across clients.
\section{Related Work}\label{sec:related-work}

\smallsection{Federated Learning with Non-IID Data}
One of the fundamental challenges in federated learning is the presence of non-IID data~\cite{kairouz2021advances}. The reasons and solutions to this challenge are being actively explored. Common solutions involve adding local regularization~\cite{li2020federated,li2021model,karimireddy2020scaffold}, improving server aggregation~\cite{wang2020federated,lin2020ensemble,wang2020tackling}, and leverage public dataset~\cite{lin2020ensemble} or synthesized features~\cite{luo2021no,zhu2021data} to calibrate models. These methods tackle more relaxed non-IID problems that assume clients have the same set of classes. As shown in our experiments, these baselines show marginal improvements over FedAvg when the clients have unaware classes. Some recent works~\cite{li2021fedrs,lin2022federated} consider the problem of clients having access to only a subset of the entire class set. For example, FedRS~\cite{li2021fedrs} addresses the case where each client only owns data from certain classes. FedPU~\cite{lin2022federated} focuses on the scenario where clients label a small portion of their datasets, and there exists unlabeled data from both positive (i.e., \textit{locally-identified} in our terminology) and negative (i.e., \textit{locally-unaware}) classes. The problem settings differ from ours. Moreover, these methods are specifically tailored for single-label classification, where the presence of one class indicates the absence or presence of other classes. When applied to our problem, they demonstrate less improvement compared to \our. 

\smallsection{Label Semantics Modeling}
In tasks where some of the label patterns cannot be directly observed from the training dataset, such as zero-shot learning~\cite{lee2018multi}, it is hard for the model to generalize to unseen classes. To deal with the problem, several methods are proposed to leverage prior knowledge such as knowledge graphs~\cite{von2021informed} or model semantic label embedding from textual information about classes~\cite{lei2015predicting,matsuki2019characterizing,wu2020multi,radford2021learning}. For example, Ba et al.~\cite{lei2015predicting} derived embedding features for classes from natural language descriptions and learned a mapping to transform text features of classes to visual image feature space. Radford et al~\cite{radford2021learning} used contrastive pretraining to jointly train an image encoder and a text encoder and predict the correct pairings of image and text caption, which helps to produce high-quality image representations. Matsuki et al~\cite{matsuki2019characterizing} and Wu et al~\cite{wu2020multi} incorporate word embeddings for zero-shot learning in human activity recognition. These methods show the potential of using semantic relationships between labels to enable predictions for classes not observed in the training set, which motivates our design of semantic label name sharing.
\section{Conclusions and Future Work}\label{sec:conclusion}
We studied the problem of federated classification with non-identical class sets. We propose the \our framework and demonstrate its use in federated learning for various applications. \our incorporates a label encoder in the backbone classification model. Semantic label learning is conducted by leveraging a domain-related corpus and shared label names. The pretrained semantic label embeddings contain the knowledge of label correlations and are used to guide the training of the data encoder. Moreover, the anchor-guided alignment enriches features for unaware classes at each client based on global class anchors and reduces the discrepancy between local distributions and global distribution. These two designs are a key to mitigating client variance in \our, which addresses the challenge of non-identical class sets. We show that \our improves the baseline algorithms for federated learning with non-IID data and achieves new state-of-the-art. 

It is worth mentioning that \our can work when the clients can only share the label IDs by assuming label names are unknown and randomly initializing the label encoder. Of course, advanced techniques like neural language models can be applied to generate and enrich the label representations, and we leave it as future work. Moreover, for future directions, we consider more general system heterogeneity where the participants have different network architectures, training processes, and tasks. We plan to extend our study to make federated learning compatible with such heterogeneity.
\section*{Acknowledgments}
This work was sponsored in part by NSF Convergence Accelerator under award OIA-2040727, NIH Bridge2AI Center Program under award 1U54HG012510-01, as well as generous gifts from Google, Adobe, and Teradata. Any opinions, findings, and conclusions or recommendations expressed herein are those of the authors and should not be interpreted as necessarily representing the views, either expressed or implied, of the U.S. Government. The U.S. Government is authorized to reproduce and distribute reprints for government purposes not withstanding any copyright annotation hereon.

\bibliographystyle{ACM-Reference-Format}
\bibliography{cited}

\newpage
\clearpage
\appendix
\section{Appendix}

\subsection{Semantic Label Embedding Pretraining}\label{sec:label-embedding-pretraining}
Before collaborative training, each client shares the natural language names of its classes with the server. The server then searches for these label names in a large text corpus related to the domain of the classification task. We count the occurrences of each label name $w_i$ in the text segments (e.g., sentences or paragraphs). We notice that the label names can be phrases that contain multiple words, and the order of the words may change in different text segments while representing the same meaning. For example, ``colon cancer'' and ``cancer of the colon'' refer to the same concept. To handle such variations, we organize the label names into sets of words and use a sliding window of length $L_w$ to scan the text segments. If the set of words in the label name is covered by the words within the sliding window, we consider the label name appears in the text segment. The length of the sliding window $L_w$ varies per label name being searched. The co-occurrence of a pair of label names $w_i$ and $w_j$ is then calculated using the point-wise mutual information (PMI): 
\[
\text{PMI}(w_i,w_j) = \log\frac{p(w_i,w_j)}{p(w_i)p(w_j)},
\]
where $p(w_i)$ and $p(w_j)$ are the individual distributions and $p(w_i,w_j)$ is the the joint distribution. The higher the $\text{PMI}(w_i,w_j)$, the stronger the association between the two label names $w_i$ and $w_j$. 

The server then learns semantic label embeddings based on the PMIs. The goal of label embedding learning is to learn a mapping function from label names to representations $\gamma: \mathcal{W} \rightarrow \mathcal{Z}$, which enforces labels with related semantic meanings to have similar representations. To achieve this, the server builds a label co-occurrence graph $G = \langle \mathbf{V}, \mathbf{E}\rangle$, where the nodes $\mathbf{V}$ represent the label names and the edges $\mathbf{E}$ represent the co-occurrence relationship between the nodes. The PMI values are zero-centered by subtracting the mean and are used as the edge weights between label names. Edges with negative weights are removed from the graph. For every source node $w \in \mathbf{V}$, we define $\mathbf{N}_s(w) \subset \mathbf{V}$ as its network neighborhood generated through simulating fixed-length random walks starting from $w$. The transition probability for random walk simulation is calculated by normalizing the edge weight. 
The objective function of label embedding learning is defined as:
\[
\mathop{\text{max}}\limits_{\gamma}\sum\limits_{w\in \mathbf{V}}(-\log Z_{w} + \sum\limits_{u\in \mathbf{N}_s(w)}\gamma(u)\cdot \gamma(w)),
\]
where $Z_{w} = \sum_{v\in \mathbf{V}} \mathop{\text{exp}}(\gamma(v)\cdot \gamma(w))$ and is approximated using negative sampling~\cite{mikolov2013distributed}. The mapping function $\gamma$ is achieved by a single hidden layer feedforward neural network and the objective is optimized using stochastic gradient descent. 

\subsection{Text Corpus}
Domain-specific raw corpora are often readily available. For example, novel books that depict daily scenes can be used for understanding human activities. Academic journals are good study resources for understanding concepts in different professional domains. We use the following corpora for applications in our experiments:
        
\noindent\textbf{BookCorpus}~\cite{zhu2015aligning} is a large collection of free novel books collected from the internet. It is a popular text corpus in the field of natural language processing and has contributed to the training of many influential language models such as BERT~\cite{devlin2018bert} and GPT~\cite{brown2020language}.

\noindent\textbf{PubMed Open-Access (OA) subset}~\cite{pubmed2003} is a text archive of journal articles and preprints in biomedical and life sciences. It has been widely used for biomedical text mining~\cite{lee2020biobert}.

\noindent\textbf{CommonCrawl (CC) News dataset}~\cite{hamborg2017news} is an English-language news corpus collecting news articles published between 2017 to 2019 from news sites worldwide.

We use PubMed-OA as the text corpus for medical code prediction, BookCorpus for behavioral context recognition and human activity recognition, and CC News for text classification. We consider a sentence as a text segment in BookCorpus, an article as a text segment in CC News, and a paragraph as a text segment in PubMed-OA. 

\subsection{Details About Datasets and Preprocessing}

\noindent\textbf{ExtraSensory dataset}~\cite{vaizman2017recognizing} collects sensory measurements from 60 participants in their natural behavior using smartphones and smartwatches. Relevant context labels of the time-series measurements are annotated. There are in total 51 classes, which can be grouped into five main categories: posture (e.g., sitting, lying down, etc.), location (e.g., gym, beach, etc.), activity (e.g., eating, sleeping, etc.), companion (with friends, with co-workers), phone position (e.g., phone in hand, etc.). Each sample is associated with 3.6 classes on average. During data preprocessing, the time-series data of each subject is partitioned into segments based on the labels, ensuring that each segment represents a consistent behavioral context.
We further organize the data into 10-minute time series. Samples with a duration of fewer than 10 minutes are padded with zeros. The features at each timestamp represent the measurements taken within one minute. The feature dimension is 225.

\noindent\textbf{MIMIC-III dataset}~\cite{johnson2016mimic} contains inpatient data from over $40,000$ patients. We follow the preprocessing instructions in \cite{mullenbach2018explainable} to derive the benchmark MIMIC-III 50-label dataset. The data for each patient consists of clinical notes that document information about the patient's hospitalization. The clinical notes have a vocabulary size of 4,896, and on average, each note has a length of 1,512 words. The dataset includes 50 classes, which can be grouped into 10 categories based on the International Classification of Diseases (ICD). Each data sample is associated with an average of 5.69 classes. The natural language label names for the MIMIC-III dataset are obtained by matching the medical codes with the ICD-9 taxonomy.

\noindent\textbf{PAMAP2}~\cite{reiss2012introducing} collects sensory measurements from 9 subjects performing 18 different physical activities (e.g., sitting, ascending stairs, etc.). The data is collected using inertial measurement units (IMU) sensors and a heart rate monitor. The feature dimension is 52. We organize the data into time series with a window size of 100.

\noindent\textbf{Reuters-21578 R8}~\cite{2007:phd-Ana-Cardoso-Cachopo} is a collection of news articles classified into 8 classes (e.g. trade, grain, earn, etc.). During preprocessing, words with at least 100 occurrences are kept, resulting in a vocabulary size of 534. Each piece of text data has a length of 66 words on average. 

\subsection{Theoretical Derivations}
\smallsection{Notations}
Let $f : \mathcal{X} \rightarrow \mathcal{Z}$ be a representation function that maps inputs to features, and $g : \mathcal{X} \rightarrow \{0, 1\}$ be a ground-truth labeling function that maps input to output space. For the global domain, denote $\mathcal{D}$ as the global distribution over the input space $\mathcal{X}$, and $\tilde{\mathcal{D}}$ as the induced global distribution over the feature latent space $\mathcal{Z}$. For the $m$-th local domain, denote $\mathcal{D}_m$ as the local distribution and $\tilde{\mathcal{D}}_m$ be the induced image of $\mathcal{D}_m$ over $\mathcal{Z}$. A hypothesis $h: \mathcal{Z} \rightarrow \{0, 1\}$ is a function that maps features to predicted labels. Let $\tilde{g}$ be the induced image of $g$ over $\mathcal{Z}$. The expected risk of hypothesis $h$ on distribution $\mathcal{D}$ is defined as follows:
\[
\mathcal{L}(h) = \mathbb{E}_{z\sim\tilde{\mathcal{D}}}[\mathbb{E}_{y\sim\tilde{g}(z)}[y \neq h(z)]].
\]

Let $\lambda_m$ denote the risk of the optimal hypothesis over hypothesis class $\mathbf{H}$ that has minimum risk on both $\mathcal{D}$ and $\mathcal{D}_m$ distributions, i.e., $\lambda_m = \mathop{\text{min}}_{h\in \mathbf{H}} (\mathcal{L}(h) + \mathcal{L}_m(h))$.

We define distance functions for measuring the divergence between two distributions with respect to the hypothesis class. First, given a feature space $\mathcal{Z}$ and a collection of measurable subsets $\mathcal{A}$ of $\mathcal{Z}$, define $\mathcal{A}$-distance between two distributions $\tilde{\mathcal{D}}$ and $\tilde{\mathcal{D}}'$ on $\mathcal{Z}$ as:
\[
d_{\mathcal{A}}(\tilde{\mathcal{D}},\tilde{\mathcal{D}}')=2\sup_{A\in\mathcal{A}}\lvert\text{Pr}_{\tilde{\mathcal{D}}}(A)-\text{Pr}_{\tilde{\mathcal{D}}'}(A)\rvert.
\]

Now, fix a particular hypothesis class $\mathbf{H}$, for any given $h\in \mathbf{H}$, define $\mathcal{Z}_h=\{z\in \mathcal{Z} \vert h(z)=1\}$ and $\mathcal{A}_\mathbf{H}=\{\mathcal{Z}_h\vert h\in \mathbf{H}\}$. Define the $\mathbf{H}$-divergence between two distributions $\tilde{\mathcal{D}}$ and $\tilde{\mathcal{D}}'$ on $\mathcal{Z}$ as:
\[
d_\mathbf{H}(\tilde{\mathcal{D}},\tilde{\mathcal{D}}')=d_{\mathcal{A}_\mathbf{H}}(\tilde{\mathcal{D}},\tilde{\mathcal{D}}').
\]

Furthermore, given a particular hypothesis class $\mathbf{H}$, define $\mathcal{A}_{\mathbf{H}\Delta \mathbf{H}}=\{\mathcal{Z}_h\Delta \mathcal{Z}_{h'}\vert h,h'\in \mathbf{H}\}$, where $\Delta$ operation is the symmetric difference in the sense of set operation. Define the $\mathbf{H}\Delta \mathbf{H}$-divergence between two distributions $\tilde{D}$ and $\tilde{D}'$ on $\mathcal{Z}$ as:
\[
d_{\mathbf{H}\Delta \mathbf{H}}(\tilde{\mathcal{D}},\tilde{\mathcal{D}}')=d_{\mathcal{A}_{\mathbf{H}\Delta \mathbf{H}}}(\tilde{\mathcal{D}},\tilde{\mathcal{D}}').\]

\noindent\begin{theorem}[Generalization Bound of Federated Learning\footnote{Proof can be found in prior work~\cite{peng2019federated,zhu2021data,lin2020ensemble}.}]
\label{theorem:generalization-FL-appendix}
Assume there are $M$ clients in a federated learning system. Let $\mathbf{H}$ be the hypothesis class with VC-dimension $d$. The global hypothesis is the aggregation of $h_m$, i.e., $h = \frac{1}{M}\sum_{m\in [M]}h_m$.
With probability at least $1-\delta$, for $\forall h \in \mathbf{H}$:
\begin{equation}
    \begin{aligned}
    \mathcal{L}(h) \leq & \frac{1}{M}\sum\limits_{m\in [M]}\hat{\mathcal{L}}_m(h_m) + \frac{1}{M}\sum\limits_{m\in [M]}[d_{\mathbf{H}\Delta \mathbf{H}}(\tilde{\mathcal{D}}_m, \tilde{\mathcal{D}})+\lambda_m]\\
    &+ \sqrt{\frac{4}{N}(d\mathop{log}\frac{2eN}{d} + \mathop{log}\frac{4M}{\delta})},
    \end{aligned}\nonumber
\end{equation}
where $\hat{\mathcal{L}}_m(h_m)$ is the empirical risk on the $m$-th client given $N$ observed samples, $e$ is the base of the natural logarithm.
\end{theorem}

\begin{corollary}[Generalization Bound of Federated Learning with Mix-up Distributions]
Let $\mathcal{D}'_m$ denote the distribution added for adapting the $m$-th client. Define the new distribution $\mathcal{D}^*_m$ to be a mixture of the original local distribution and the adaptation distribution, i.e., $\mathcal{D}^*_m = \alpha\mathcal{D}_m + (1-\alpha)\mathcal{D}_m'$, where $\alpha \in [0, 1]$ is the weight of the original distribution decided by the number of empirical samples added. Let $\mathbf{H}$ be the hypothesis class with VC-dimension $d$. The global hypothesis is the aggregation of $h_m$, i.e., $h = \frac{1}{M}\sum_{m\in [M]}h_m$. With probability at least $1-\delta$, for $\forall h \in \mathbf{H}$:
\begin{equation}
\resizebox{\linewidth}{!}{$
    \begin{aligned}
    \mathcal{L}(h) \leq & \frac{1}{M}\sum\limits_{m\in [M]}\hat{\mathcal{L}}_m(h_m)\\
    &+ \frac{1}{M}\sum\limits_{m\in[M]}[\alpha d_{\mathbf{H}\Delta \mathbf{H}}(\tilde{\mathcal{D}}_m, \tilde{\mathcal{D}}) + (1-\alpha)d_{\mathbf{H}\Delta \mathbf{H}}(\tilde{\mathcal{D}}'_m, \tilde{\mathcal{D}}) + \lambda_m]\\
    &+ \sqrt{\frac{4}{N^*}(d\mathop{log}\frac{2eN^*}{d} + \mathop{log}\frac{4M}{\delta})},
    \end{aligned}\nonumber
$}
\end{equation}
where $\hat{\mathcal{L}}_m(h_m)$ is the empirical risk on the $m$-th client given $N^*$ ($N^* > N$) observed samples, $e$ is the base of the natural logarithm. 

\end{corollary}

\begin{proof}
Apply Theorem~\ref{theorem:generalization-FL-appendix} to the mix-up distribution $\mathcal{D}^*_m$, we have that with probability at least $1-\delta$, for $\forall h \in \mathbf{H}$:
\begin{equation}
    \begin{aligned}
    \mathcal{L}(h) \leq & \frac{1}{M}\sum\limits_{m\in [M]}\hat{\mathcal{L}}_m(h_m) + \frac{1}{M}\sum\limits_{m\in [M]}[d_{\mathbf{H}\Delta \mathbf{H}}(\tilde{\mathcal{D}}^*_m, \tilde{\mathcal{D}})+\lambda_m]\\
    &+ \sqrt{\frac{4}{N^*_m}(d\mathop{log}\frac{2eN^*_m}{d} + \mathop{log}\frac{4M}{\delta})}.
    \end{aligned}\nonumber
\end{equation}

We derive the upper bound of $\mathbf{H}\Delta \mathbf{H}$-divergence between the mix-up distribution $\mathcal{D}^*_m$ and the global distribution $\mathcal{D}$ as follows:
\begin{equation}
    \begin{aligned}
    & d_{\mathbf{H}\Delta \mathbf{H}}(\tilde{\mathcal{D}}^*_m, \tilde{\mathcal{D}})
    = 2 \mathop{\text{sup}}\limits_{A\in \mathcal{A}_{\mathbf{H}\Delta \mathbf{H}}}\lvert \text{Pr}_{\tilde{\mathcal{D}}^*_m}(A) - \text{Pr}_{\tilde{\mathcal{D}}}(A)\rvert\\
    &= 2 \mathop{\text{sup}}\limits_{A\in \mathcal{A}_{\mathbf{H}\Delta \mathbf{H}}}\lvert \text{Pr}_{\alpha\tilde{\mathcal{D}}_m+(1-\alpha)\tilde{\mathcal{D}}_m'}(A) - \text{Pr}_{\tilde{\mathcal{D}}}(A)\rvert\\
    &= 2 \mathop{\text{sup}}\limits_{A\in \mathcal{A}_{\mathbf{H}\Delta \mathbf{H}}}\lvert \alpha\text{Pr}_{\tilde{\mathcal{D}}_m}(A) + (1-\alpha)\text{Pr}_{\tilde{\mathcal{D}}_m'}(A) - \text{Pr}_{\tilde{\mathcal{D}}}(A)\rvert\\
    &\leq 2 \alpha\mathop{\text{sup}}\limits_{A\in \mathcal{A}_{\mathbf{H}\Delta \mathbf{H}}}\lvert \text{Pr}_{\tilde{\mathcal{D}}_m}(A) - \text{Pr}_{\tilde{\mathcal{D}}}(A)\rvert\\
    &\quad + 2(1-\alpha)\mathop{\text{sup}}\limits_{A\in \mathcal{A}_{\mathbf{H}\Delta \mathbf{H}}}\lvert\text{Pr}_{\tilde{\mathcal{D}}_m'}(A) - \text{Pr}_{\tilde{\mathcal{D}}}(A)\rvert\\
    &=\alpha d_{\mathbf{H}\Delta \mathbf{H}}(\tilde{\mathcal{D}}_m, \tilde{\mathcal{D}}) + (1-\alpha)d_{\mathbf{H}\Delta \mathbf{H}}(\tilde{\mathcal{D}}'_m, \tilde{\mathcal{D}}).
    \end{aligned}\nonumber
\end{equation}
The inequality is derived using the triangle inequality. Replace the $\mathcal{A}$-distance $d_{\mathbf{H}\Delta \mathbf{H}}(\tilde{\mathcal{D}}^*_m, \tilde{\mathcal{D}})$ with its upper bound derived above, we get the upper bound of $\mathcal{L}(h)$.

If $d_{\mathbf{H}\Delta \mathbf{H}}(\tilde{\mathcal{D}}'_m, \tilde{\mathcal{D}}) < d_{\mathbf{H}\Delta \mathbf{H}}(\tilde{\mathcal{D}}_m, \tilde{\mathcal{D}})$, then we have:

\begin{equation}
    \begin{aligned}
    d_{\mathbf{H}\Delta \mathbf{H}}(\tilde{\mathcal{D}}^*_m, \tilde{\mathcal{D}}) 
    &\le\alpha d_{\mathbf{H}\Delta \mathbf{H}}(\tilde{\mathcal{D}}_m, \tilde{\mathcal{D}}) + (1-\alpha)d_{\mathbf{H}\Delta \mathbf{H}}(\tilde{\mathcal{D}}'_m, \tilde{\mathcal{D}})\\
    &< d_{\mathbf{H}\Delta \mathbf{H}}(\tilde{\mathcal{D}}_m, \tilde{\mathcal{D}}).
    \end{aligned}\nonumber
\end{equation}

Furthermore, with added empirical samples from $D'_m$, we have $N^* > N$, and:
\begin{equation}
    \sqrt{\frac{4}{N^*}(d\mathop{log}\frac{2eN^*}{d} + \mathop{log}\frac{4M}{\delta})} < \sqrt{\frac{4}{N}(d\mathop{log}\frac{2eN}{d} + \mathop{log}\frac{4M}{\delta})}.
    \nonumber
\end{equation}

Therefore, the upper bound of the expected risk with the mix-up distribution is lowered.
\end{proof}

\end{document}